\documentclass[10pt,logo,copyright]{xhumanoid-giga-report}
\usepackage[authoryear,sort&compress,round]{natbib}

\usepackage[utf8]{inputenc} %
\usepackage[T1]{fontenc}    %

\usepackage{parskip}        %
\usepackage{url}            %
\usepackage{booktabs}       %
\usepackage{amsfonts}       %
\usepackage{nicefrac}       %
\usepackage{microtype}      %
\usepackage{xcolor}         %
\usepackage[dvipsnames]{xcolor} %
\usepackage{graphicx}
\usepackage{animate}        %
\usepackage{subcaption}
\usepackage{tabularx}
\usepackage{makecell}
\usepackage{adjustbox}
\usepackage{setspace}
\usepackage{todonotes}
\usepackage{colortbl}
\usepackage{wrapfig}

\newcolumntype{M}[1]{>{\centering\arraybackslash}m{#1}}
\usepackage{float}
\usepackage{tikz}
\usetikzlibrary{positioning,shapes,arrows}
\usepackage{amsmath,amsfonts,bm, bbm,leftindex}
\usepackage{multirow}
\usepackage{comment}
\usepackage{gensymb}
\usepackage{lipsum}
\usetikzlibrary{arrows.meta, positioning, fit}
\usepackage[para]{threeparttable}
\usepackage{tikz}
\usetikzlibrary{tikzmark}

\def\eqref#1{equation~\ref{#1}}

\def\1{\bm{1}}

\DeclareMathAlphabet{\mathsfit}{\encodingdefault}{\sfdefault}{m}{sl}
\SetMathAlphabet{\mathsfit}{bold}{\encodingdefault}{\sfdefault}{bx}{n}

\makeatletter
\let\save@mathaccent\mathaccent
\newcommand*\if@single[3]{%
  \setbox0\hbox{${\mathaccent"0362{#1}}^H$}%
  \setbox2\hbox{${\mathaccent"0362{\kern0pt#1}}^H$}%
  \ifdim\ht0=\ht2 #3\else #2\fi
  }
\newcommand*\rel@kern[1]{\kern#1\dimexpr\macc@kerna}
\newcommand*\widebar[1]{\@ifnextchar^{{\wide@bar{#1}{0}}}{\wide@bar{#1}{1}}}
\newcommand*\wide@bar[2]{\if@single{#1}{\wide@bar@{#1}{#2}{1}}{\wide@bar@{#1}{#2}{2}}}
\newcommand*\wide@bar@[3]{%
  \begingroup
  \def\mathaccent##1##2{%
    \let\mathaccent\save@mathaccent
    \if#32 \let\macc@nucleus\first@char \fi
    \setbox\z@\hbox{$\macc@style{\macc@nucleus}_{}$}%
    \setbox\tw@\hbox{$\macc@style{\macc@nucleus}{}_{}$}%
    \dimen@\wd\tw@
    \advance\dimen@-\wd\z@
    \divide\dimen@ 3
    \@tempdima\wd\tw@
    \advance\@tempdima-\scriptspace
    \divide\@tempdima 10
    \advance\dimen@-\@tempdima
    \ifdim\dimen@>\z@ \dimen@0pt\fi
    \rel@kern{0.6}\kern-\dimen@
    \if#31
      \overline{\rel@kern{-0.6}\kern\dimen@\macc@nucleus\rel@kern{0.4}\kern\dimen@}%
      \advance\dimen@0.4\dimexpr\macc@kerna
      \let\final@kern#2%
      \ifdim\dimen@<\z@ \let\final@kern1\fi
      \if\final@kern1 \kern-\dimen@\fi
    \else
      \overline{\rel@kern{-0.6}\kern\dimen@#1}%
    \fi
  }%
  \macc@depth\@ne
  \let\math@bgroup\@empty \let\math@egroup\macc@set@skewchar
  \mathsurround\z@ \frozen@everymath{\mathgroup\macc@group\relax}%
  \macc@set@skewchar\relax
  \let\mathaccentV\macc@nested@a
  \if#31
    \macc@nested@a\relax111{#1}%
  \else
    \def\gobble@till@marker##1\endmarker{}%
    \futurelet\first@char\gobble@till@marker#1\endmarker
    \ifcat\noexpand\first@char A\else
      \def\first@char{}%
    \fi
    \macc@nested@a\relax111{\first@char}%
  \fi
  \endgroup
}
\makeatother

\definecolor{darkred}{rgb}{0.7, 0.0, 0.0}

\usepackage{amsmath} 

\usepackage{pifont}

\usepackage[nameinlink]{cleveref}
\crefname{equation}{Eq.}{Eqs.}
\crefname{figure}{Fig.}{Figs.}
\crefname{section}{Sec.}{Sec.}
\crefname{appendix}{App.}{App.}
\crefname{table}{Tab.}{Tabs.}
\crefname{algorithm}{Algo}{Algo}
\crefname{thm}{Thm}{Thm}
\Crefname{thm}{Thm}{Thm}
\crefname{prop}{Prop}{Prop}

\newcommand{\crefnames}[3]{%
  \@for\next:=#1\do{%
    \expandafter\crefname\expandafter{\next}{#2}{#3}%
  }%
}

\title{Humanoid Occupancy: Enabling A Generalized Multimodal Occupancy Perception System on Humanoid Robots}

\author{Wei Cui$^*$$^1$, Haoyu Wang$^*$$^2$, Wenkang Qin$^2$, Yijie Guo$^1$, Gang Han$^1$, Wen Zhao$^1$, Jiahang Cao$^1$, Zhang Zhang$^1$, Jiaru Zhong$^1$, Jingkai Sun$^1$, Pihai Sun$^1$, Shuai Shi$^1$, Botuo Jiang$^1$, Jiahao Ma$^1$, Jiaxu Wang$^1$, Hao Cheng$^1$, Zhichao Liu$^2$, Yang Wang$^2$, Zheng Zhu$^2$, Guan Huang$^2$, Jian Tang$^\dagger$$^1$ and Qiang Zhang$^\dagger$$^\Delta$$^1$ \\
$^*$ Equal Contributors,  $^\dagger$ Corresponding Authors, $^\Delta$ Project and Technical Leader \\
$^1$ X-Humanoid,  $^2$ GigaAI \\
Project Page: \href{https://humanoid-occupancy.github.io}{https://humanoid-occupancy.github.io}}

\begin{abstract}
Humanoid robot technology is advancing rapidly, with manufacturers introducing diverse heterogeneous visual perception modules tailored to specific scenarios. Among various perception paradigms, occupancy-based representation has become widely recognized as particularly suitable for humanoid robots, as it provides both rich semantic and 3D geometric information essential for comprehensive environmental understanding.

In this work, we present Humanoid Occupancy, a generalized multimodal occupancy perception system that integrates hardware and software components, data acquisition devices, and a dedicated annotation pipeline. Our framework employs advanced multi-modal fusion techniques to generate grid-based occupancy outputs encoding both occupancy status and semantic labels, thereby enabling holistic environmental understanding for downstream tasks such as task planning and navigation. To address the unique challenges of humanoid robots, we overcome issues such as kinematic interference and occlusion, and establish an effective sensor layout strategy. Furthermore, we have developed the first panoramic occupancy dataset specifically for humanoid robots, offering a valuable benchmark and resource for future research and development in this domain. The network architecture incorporates multi-modal feature fusion and temporal information integration to ensure robust perception. Overall, Humanoid Occupancy delivers effective environmental perception for humanoid robots and establishes a technical foundation for standardizing universal visual modules, paving the way for the widespread deployment of humanoid robots in complex real-world scenarios.
\end{abstract}

\begin{document}


\maketitle
\vspace{-0.2in}
\begin{center}
    \centering
    \captionsetup{type=figure}
    \includegraphics[width=0.99\linewidth]{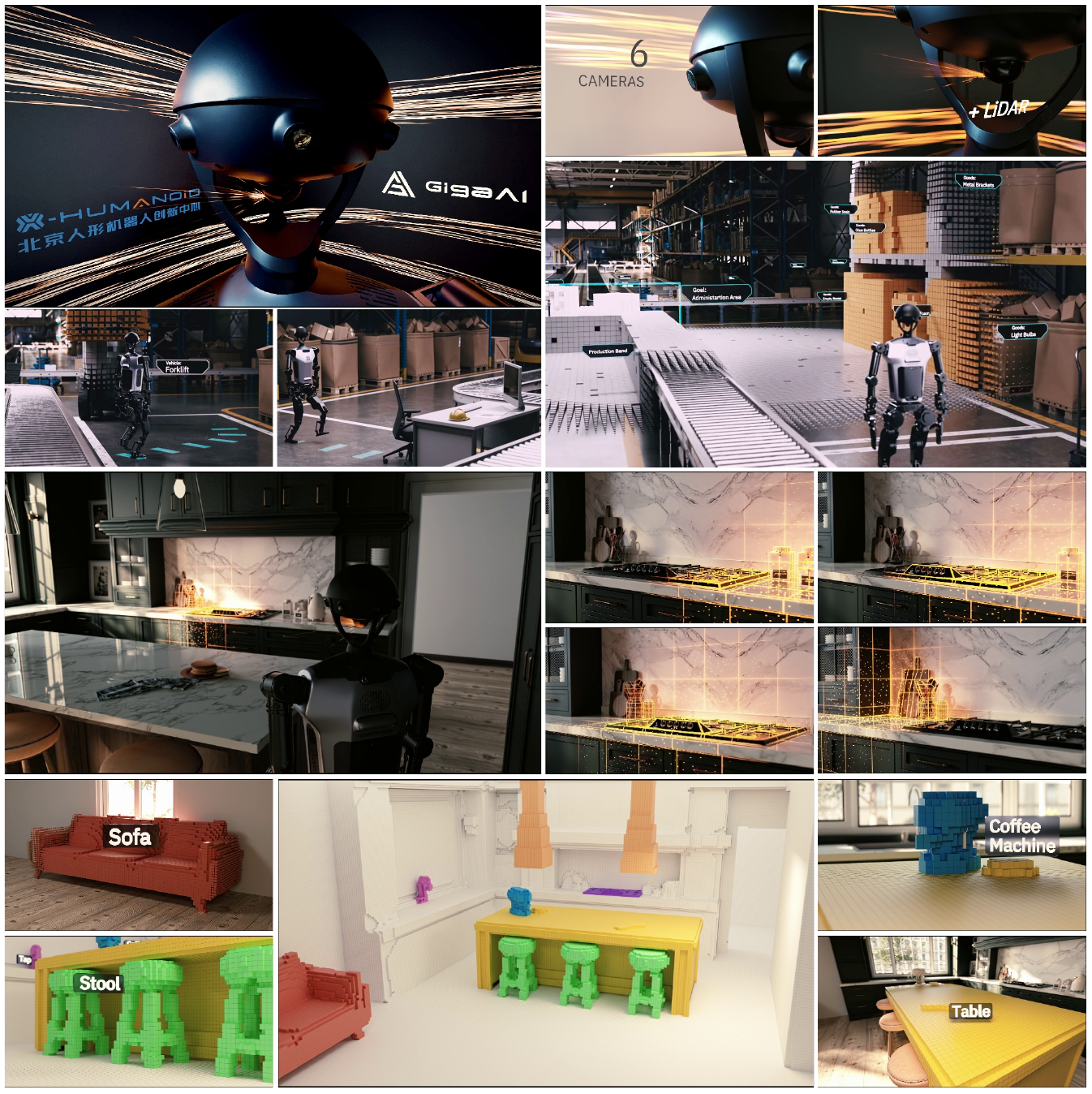}
    \caption{\textbf{Schematic diagram of the Humanoid Occupancy system.}}
    \label{fig:teaser}
\end{center}
\vspace{-0.05in}
\clearpage
\abscontent
\section{Introduction}

Humanoid robots are considered the most promising general-purpose robots for integration into human environments due to their human-like structure and movement~\citep{hirai1998development}. Despite rapid progress in AI and robotics, their visual perception systems still face key challenges~\citep{bonci2021human}. These include designing optimal multimodal sensor layouts for complex environments, efficiently collecting and annotating large-scale diverse datasets, and achieving generality and standardization in visual system design. 

From a task-driven perspective, humanoid robots face three core tasks: \textcolor{orange}{manipulation, locomotion, and navigation}~\citep{gu2025humanoid}. Manipulation tasks require rich texture and geometric information for precise object recognition and handling; locomotion tasks emphasize terrain geometry perception to ensure stable walking in complex environments; navigation tasks demand a global understanding of environmental semantics and spatial geometry for efficient path planning and autonomous navigation. These diverse requirements present significant challenges for the design of perception systems~\citep{roychoudhury2023perception, disalvo2002all}.

In this work, we focus on enhancing the environmental perception and navigation capabilities of humanoid robots. Environmental perception forms the basis for executing complex tasks, while navigation remains a core challenge in humanoid robotics, with notable advances in areas such as autonomous driving and aerial robotics. However, the unique structural characteristics of humanoid robots introduce challenges such as kinematic interference, data scarcity, and limited representational capacity, which continue to restrict their effective deployment in real-world environments.

We propose the Humanoid Occupancy system (\Cref{fig:teaser}) to provide humanoid robots with a unified, efficient, and information-rich environmental perception capability. The choice of occupancy as the core visual representation paradigm is motivated by several considerations. First, occupancy can directly encode the occupancy status of each spatial unit in the environment in a voxel or grid format, capturing not only the distribution on the 2D plane but also detailed structural and semantic attributes along the vertical dimension. This makes it far superior to traditional representations such as Bird’s Eye View (BEV)~\citep{philion2020lift}, which typically focus only on ground projections and fail to reflect vertical structures and high-level semantic information. Second, occupancy representation~\citep{elfes2013occupancy} is naturally suited for multimodal data fusion, allowing information from RGB, depth, LiDAR, and other sensors to be unified within a spatial grid for comprehensive environmental understanding. Compared to other 3D representations such as point clouds~\citep{rusu20113d} or meshes, occupancy offers greater generality and extensibility in data structure, semantic annotation, and downstream task interfaces, facilitating large-scale data processing and model training. Moreover, the dense spatial distribution output by occupancy directly supports path planning, obstacle avoidance, and manipulation tasks, significantly enhancing the adaptability of humanoid robots in complex environments. Therefore, occupancy representation provides a solid foundation for efficient perception across multiple tasks and scenarios.

In terms of system implementation, we adopt a standard three-stage visual system architecture, comprising hardware design, dataset construction, and a multimodal fusion network. We propose an innovative sensor layout strategy tailored for humanoid robots, effectively mitigating perception blind spots caused by structural interference. Additionally, we have constructed the first panoramic occupancy dataset specifically for humanoid robots, providing a valuable benchmark and resource for future research and applications. Finally, we have integrated and tested the system on the Tienkung humanoid robot platform, and experimental results demonstrate its superior environmental perception and navigation performance in complex environments.
\section{Preliminaries and Related Works}

\subsection{Perception System of Humanoid Robots}
The development of visual perception systems for humanoid robots has evolved from single-sensor configurations to multimodal fusion architectures~\citep{IEEE_Spectrum_01}. Early systems primarily relied on monocular or stereo cameras~\citep{hirose2007honda, shigemi2017asimo, mutlu2006perceptions, cheng2024open, fu2024humanplus, lu2024mobile, ben2025homie, peng2025lovon}, but their limited perception range and depth sensing capabilities made it difficult to meet the demands of complex scenarios. With advances in sensor technology, RGB-D cameras~\citep{zhang2025distillation, sun2025trinity, sun2025learning, ze2024generalizable, long2411learning, duan2024learning, zhuang2024humanoid}, LiDAR~\citep{wang2025omni, zhang2025learning}, panoramic cameras~\citep{zhang2025humanoidpano}, and other sensor types~\citep{kim2024armor, he2025attention} have been increasingly integrated into robotic platforms, laying the foundation for more comprehensive environmental perception in humanoid robots.

Current mainstream visual perception systems for humanoid robots can be categorized into the following technical paradigms:

\begin{itemize}
\item \textbf{Vision-Based Perception Systems}

These systems are centered around camera arrays and leverage computer vision and deep learning algorithms for object detection, semantic segmentation, and 3D reconstruction. A typical example is Boston Dynamics’ Atlas robot~\citep{IEEE_Spectrum_01}, which is equipped with multiple camera modules for environmental understanding and navigation. In recent years, breakthroughs in Transformer architectures and self-supervised learning have significantly improved the generalization capabilities of these systems in object recognition and scene understanding tasks~\citep{kirillov2023segment, carion2020end, radford2021learning}.

\item \textbf{Multimodal Fusion Perception Systems}

By fusing RGB images, depth data, LiDAR point clouds, and other sources, these systems exploit the complementarity of different sensors to enhance perception accuracy and robustness. For example, Agility Robotics’ Digit robot~\citep{agilityrobotics} employs a fusion scheme of LiDAR and vision, enabling stable locomotion and obstacle detection in complex terrains. Multimodal fusion not only improves the perception of dynamic obstacles and textureless regions but also enhances adaptability under varying lighting conditions. Current research focuses on deep learning-based multimodal feature fusion networks~\citep{rudin2025parkour}, which effectively integrate heterogeneous sensor information to further optimize the accuracy of environmental modeling.

\item \textbf{End-to-End Perception and Decision-Making Systems}

These frameworks utilize deep neural networks to integrate perception, understanding, and decision-making, directly outputting motion control commands~\citep{luo2024pie, wang2025omni}. While they demonstrate strong adaptability in specific tasks, they require large-scale datasets and significant computational resources. 
Recently, reinforcement learning and imitation learning have been widely applied in end-to-end systems, advancing the intelligence of humanoid robots in autonomous navigation and manipulation tasks.

\item \textbf{Spatial Representation-Based Perception Systems}

Spatial representation methods such as occupancy grids, point clouds, and meshes have become core technologies for providing structured environmental information to downstream tasks (e.g., navigation, manipulation)~\citep{zhang2025occupancy, zhang2025roboocc}. Among them, occupancy representations~\citep{sima2023scene}, which encode both rich semantic and geometric information, have emerged as a key direction for enhancing environmental understanding. These methods support 3D map construction and dynamic scene modeling, and provide efficient interfaces for path planning and obstacle avoidance. Mainstream techniques include voxel-based dense reconstruction and neural implicit field-based scene modeling, which have greatly advanced high-precision perception of complex environments.
\end{itemize}

Notably, the emergence of novel spatial representation methods such as 3D Gaussian Splatting (3DGS)~\citep{kerbl20233d, wang2024query, wang2024reinforcement} and Neural Radiance Fields (NeRF)~\citep{mildenhall2021nerf, ze2023gnfactor} has further improved the expressive power and rendering efficiency of spatial representations. 3DGS achieves fast rendering and reconstruction of large-scale scenes through efficient Gaussian distribution modeling, while NeRF employs neural networks for implicit modeling, enabling high-quality 3D reconstruction and novel view synthesis. These methods offer new technical pathways for real-time perception and interaction in complex environments, expanding the application boundaries of spatial representations in robotics.

With advances in computing hardware and algorithms, real-time perception and large-scale data processing capabilities have improved significantly. The emergence of open-source platforms such as ROS~\citep{macenski2022robot, macenski2023impact, ros} and Open3D~\citep{zhou2018open3d}, as well as standardized datasets, has accelerated the iteration and cross-platform application of visual perception systems. Current research also focuses on key technologies such as sensor self-calibration, self-supervised data annotation, and cross-domain generalization to further enhance the practicality and robustness of these systems. Overall, visual perception systems for humanoid robots are evolving toward multimodal fusion, optimized spatial representation, and integrated intelligent decision-making. How to efficiently arrange sensors within limited space and achieve effective data fusion and representation has become a core issue for improving system performance, laying a solid foundation for the introduction of occupancy representations and subsequent system design.

Building on these paradigms, this work proposes a novel occupancy perception framework for humanoid robots, which integrates multimodal sensors through a front-end fusion architecture. This framework adopts a annotation pipeline and incorporates temporal feature fusion to enhance dynamic scene understanding, aiming to address key challenges in practical applications.

\subsection{Occupancy Perception}

Occupancy perception has emerged as a key paradigm in 3D scene understanding, especially in autonomous driving, where occupancy networks and grid-based representations enable robust semantic mapping and navigation. These advances have demonstrated the power of dense spatial reasoning, multimodal fusion, and real-time inference for complex environments. However, the application of occupancy-based methods in humanoid robotics remains limited, despite their unique suitability for human-centric, dynamic scenarios.

For humanoid robots, occupancy perception offers distinct advantages: it supports fine-grained spatial reasoning in near-field environments, which is essential for precise foot placement, manipulation guidance, and safe navigation in cluttered spaces. Unlike vehicles, humanoid robots require high-resolution occupancy maps within short ranges and must address frequent occlusions and articulated motions. This creates new challenges for sensor layout, data annotation, and multimodal fusion, demanding solutions tailored to the robot’s kinematics and operational context.

Recent research has begun to adapt occupancy networks for robotic platforms, integrating RGB, depth, and LiDAR sensors to construct unified occupancy grids, developing annotation pipelines for dynamic scenes, and designing fusion networks that leverage temporal and semantic cues~\citep{wu2024embodiedoccembodied3doccupancy, wang2025embodiedoccboostingembodied3d}. These efforts aim to provide holistic environmental understanding for manipulation, obstacle avoidance, and autonomous navigation~\citep{zhang2025humanoidpano}. Nevertheless, robust multimodal fusion, real-time inference, and scalable dataset construction remain open challenges, compounded by the lack of standardized benchmarks for humanoid robots.

We can broadly categorize prior work on occupancy according to our requirements into three main types:

\begin{itemize}
\item \textbf{Occupancy Perception Datasets}

Recent advances in 3D occupancy prediction heavily rely on large-scale annotated datasets.
OpenOccupancy~\citep{wang2023openoccupancy}  pioneers the first benchmark for surround-view occupancy perception, providing dense voxel-level annotations for LiDAR and camera data.
Similarly, Occ3D~\citep{tian2023occ3d} introduces a label generation pipeline that produces dense, visibility-aware labels for any given scene. 
However, these datasets focus on autonomous driving scenarios with long-range perception (50-200m) and meter-level resolution, neglecting two critical aspects of humanoid robotics:
(1) Near-field high-resolution demands – requiring decimeter-level accuracy in 10m range for foot placement;
(2) Unstructured human interactions – needing dense annotations for articulated motions and dynamic scenarios.

\item \textbf{Vision-Based Occupancy Prediction}

Pure camera-based approaches have evolved rapidly by leveraging bird's-eye-view (BEV) representations~\citep{ma2024vision}.
TPVFormer~\citep{huang2023tri} proposes a tri-perspective view decomposition for efficient 3D scene modeling from 2D images.
FB-Occ~\citep{li2023fb} introduces a forward-backward projection mechanism to enhance occlusion reasoning, while FlashOcc~\citep{yu2023flashocc} achieves real-time performance through compressed volumetric features. 
Complementary works like VoxFormer~\citep{li2023voxformer} and OccFormer~\citep{zhang2023occformer} demonstrate transformer-based voxel prediction advancements.
SparseOcc~\citep{tang2024sparseocc} utilizes mask-guide sparse sampling to enable sparse queries to optimize efficiency.
Despite their progress, these methods often struggle with depth ambiguity in cluttered environments.

\item \textbf{Multimodal Fusion for Occupancy}

Multisensor fusion methods combine complementary sensor strengths.
BEVFusion~\citep{liang2022bevfusion, liu2022bevfusion} unifies LiDAR and camera features through late fusion in BEV space, achieving superior performance on nuScenes~\citep{caesar2020nuscenes}.
OccFusion~\citep{zhang2024occfusion, ming2024occfusion} extends this with voxel-based cross-modal attention, enabling dense occupancy prediction from asynchronous sensor streams.
These frameworks establish strong baselines for multimodal 3D perception.

\end{itemize}

\section{Humanoid Occupancy}

\subsection{System Overview}

The design of perception systems for humanoid robots must simultaneously address the requirements of manipulation, locomotion, and navigation. These three core tasks often impose conflicting constraints, necessitating careful trade-offs in sensor selection and placement. For example, mounting the sensor suite on the chest is advantageous for navigation and locomotion, providing a stable and unobstructed field of view for mapping and obstacle avoidance. However, this configuration can lead to occlusions and interference during tabletop manipulation or when handling objects such as boxes, limiting the robot's operational versatility.

The development of visual sensor technology for humanoid robots remains an ongoing process, with many sensors imposing strict structural requirements. For instance, LiDAR sensors are typically unsuitable for installation on moving, non-rigid joints unless highly accurate IMU devices are available to provide precise pose and location. This constraint further complicates sensor layout, as designers must balance the need for comprehensive environmental coverage with the mechanical limitations of the robot's kinematic structure.
\begin{wrapfigure}{r}{0.4\textwidth}
    \centering
    \includegraphics[width=1.\linewidth]{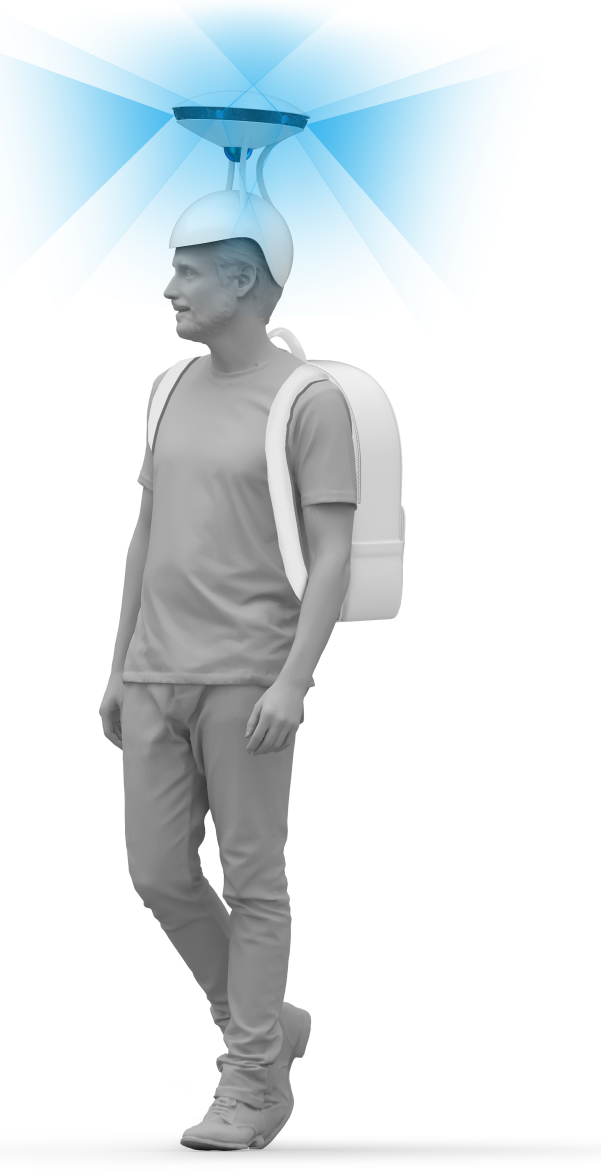}
    \caption{\textbf{Equipment and Schematic Diagram of Data Acquisition Process.}}
    \label{fig:data_collection}
\end{wrapfigure}

In this work, we implement the Humanoid Occupancy concept system on the Tienkung humanoid robot. The system features a modular, selectable RGB-D camera with two degrees of freedom (pitch and yaw), supporting both manipulation and terrain perception tasks. Additionally, 6 cameras and one LiDAR sensor are deployed to enable robust mapping and semantic understanding of the surrounding environment. This configuration is designed to maximize the robot's perceptual capabilities across diverse operational scenarios, while minimizing interference and ensuring adaptability for future sensor upgrades.

\begin{figure}[t]
    \centering
    \includegraphics[width=0.98\linewidth]{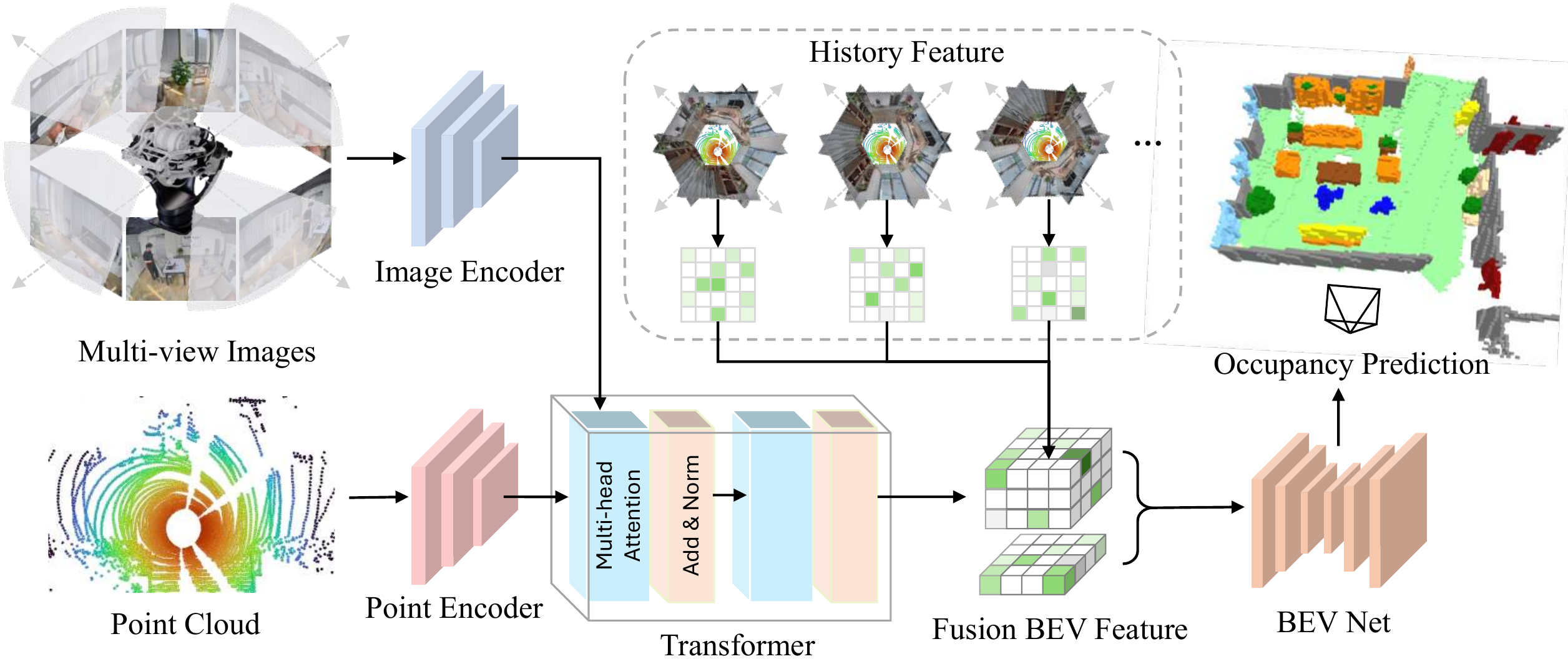}
    \caption{\textbf{Overview of our proposed HumaniodOcc.} The proposed network processes image and LiDAR inputs through separate encoders to extract features. The image features and LiDAR features are then fused via cross-attention mechanisms, enabling adaptive interaction between visual and geometric cues. The fused features undergo temporal fusion to aggregate sequential information. Finally, the network predicts the 3D occupancy grid using a BEV encoder and decoder head. More details in~\Cref{chap:fusion_network}.}
    \label{fig:model_arch}
\end{figure}

\subsection{Sensor Layout and Data Acquisition}
\label{chap:sensor_layout}

\begin{figure}[]
    \centering
    \begin{subfigure}[b]{0.45\textwidth}
        \centering
        \includegraphics[width=\textwidth]{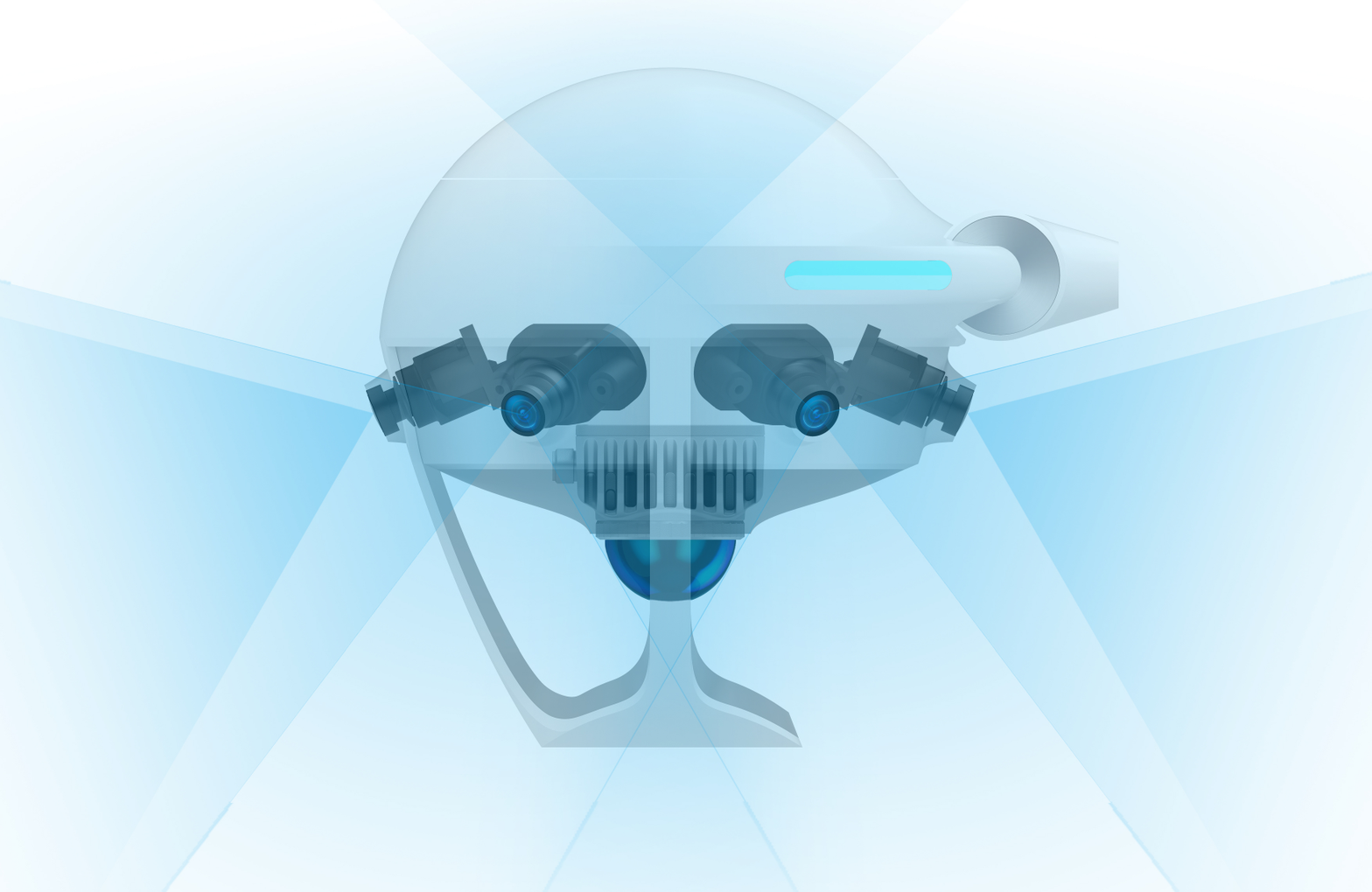}
        \caption{Side View}
    \end{subfigure}
    \hfill
    \begin{subfigure}[b]{0.45\textwidth}
        \centering
        \includegraphics[width=\textwidth]{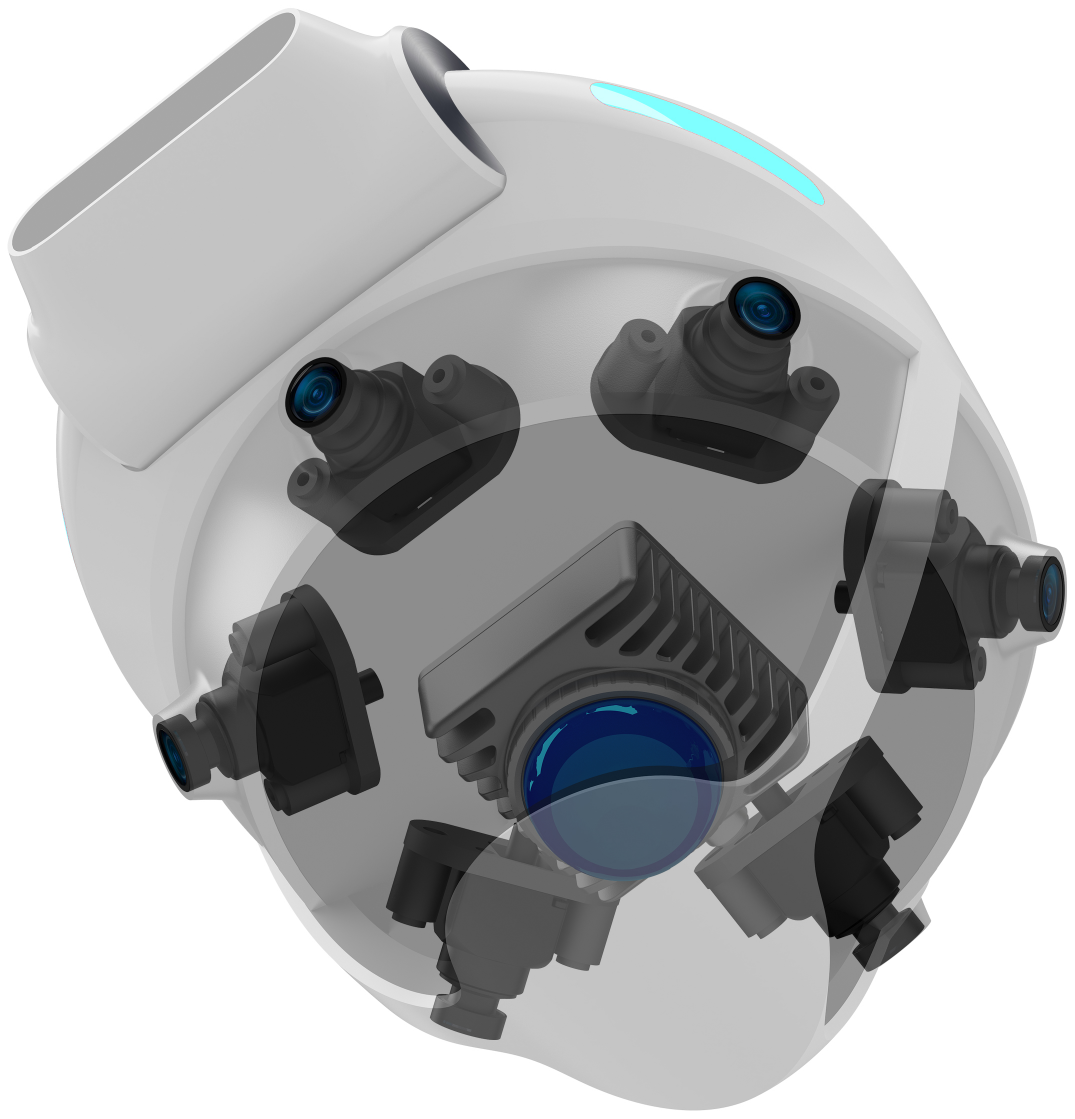}
        \caption{Isometric View}
    \end{subfigure}
    \caption{\textbf{Structural visualization of humanoid occupancy hardware.} We equipped the humanoid robot head with 6 cameras and a lidar to acquire sensory information. More details in \Cref{chap:sensor_layout}}
    \label{fig:head_structure}
\end{figure}

\noindent\textbf{Sensor Layout}
Our sensor consists of 6 cameras and a LiDAR.
The 6 cameras use standard RGB sensors, arranged in a way that one is arranged in the front and back, and two are arranged on each side.
The horizontal FOV of the camera is 118 degrees and the vertical FOV is 92 degrees.
The LiDAR uses a 40-line 360-degree omnidirectional LiDAR with a vertical FOV of 59 degrees.
The final hardware device structure can be referred to \Cref{fig:head_structure}.

\noindent\textbf{Data Acquisition}
When collecting data for autonomous driving perception models, vehicles are usually used to collect data directly on the road.
But when it comes to humanoid occupancy data acquisition, it is hard to use humanoid robotics to collect data in the environments, due to the extremely high cost and considerable difficulty.
we use a wearable device to overcome the above challenges.
Our device is constructed using the same sensor configuration as that used on humanoid robots as shown in \Cref{fig:data_collection}.
So human data collectors can wear the device directly on their heads to collect data in the environment.
This collection method greatly reduces our collection cost and difficulty.

In order to ensure that the distribution of collected data is as close as possible to the distribution of real robot data, we require the height of human data collectors to be around 160 cm.
This ensures that the height of the sensors worn by human collectors is almost the same as the height of the sensors finally installed on the humanoid robot.
To prevent the head shaking of human collectors during collection, causing the sensor to lose horizontal stability, we also added a neck stabilizer to the collection equipment.

\subsection{Annotation Pipeline}
\label{chap:annotation}
\begin{figure}[]
    \centering
    \includegraphics[width=0.8\linewidth]{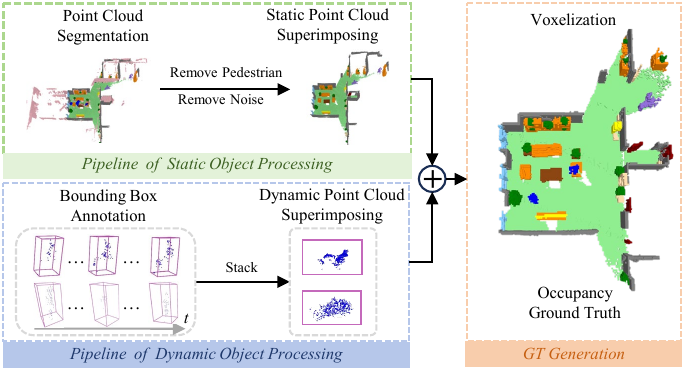}
    \caption{\textbf{The occupancy generation pipeline.} The ground truth generation pipeline begins with 3D bounding box annotation and semantic segmentation labeling. Multi-frame LiDAR scans are then processed by separately stitching dynamic objects through bounding box alignment and accumulating static scenes via motion alignment. Finally, the merged points are voxelized to produce the 3D occupancy grid with both occupancy status and semantic labels.
    More details in \Cref{chap:annotation}.}
    \label{fig:occ_generation_pipeline}
\end{figure}
We divide the collected data into three categories according to the scenes, namely home scenes, industrial scenes and outdoor scenes.
In the three different scenes, we define different point-wise semantic categories to be annotated.
We annotate the bounding boxes of dynamic objects, including pedestrians, cyclists and vehicles.
Pedestrians are non-rigid targets, and sometimes other object points are included when they are annotated using bounding boxes.
To solve this problem, we divide pedestrians into special posture pedestrians, that is, pedestrians that cannot be represented by bounding boxes, and ordinary posture pedestrians, that is, pedestrians that can be represented by bounding boxes.
Ordinary posture pedestrians only require bounding box annotation, while special posture pedestrians require not only bounding box annotation but also point-by-point annotation within the bounding box to distinguish pedestrians from other objects.
\begin{table}[ht]
\centering
\caption{\textbf{Point semantic categories.} We used different sets of semantic labels in three different scenarios.}
\label{table:annotation_semantics}
\begin{tabular}{|cccccccc}
\hline
\multicolumn{8}{|c|}{\cellcolor[HTML]{C0C0C0}{\color[HTML]{000000} Home scenes}} \\ \hline
\multicolumn{1}{|c|}{1} &
  \multicolumn{1}{c|}{2} &
  \multicolumn{1}{c|}{3} &
  \multicolumn{1}{c|}{4} &
  \multicolumn{1}{c|}{5} &
  \multicolumn{1}{c|}{6} &
  \multicolumn{1}{c|}{7} &
  \multicolumn{1}{c|}{8} \\ \hline
\multicolumn{1}{|c|}{pedestrian} &
  \multicolumn{1}{c|}{robot} &
  \multicolumn{1}{c|}{chair} &
  \multicolumn{1}{c|}{table} &
  \multicolumn{1}{c|}{floor} &
  \multicolumn{1}{c|}{wall} &
  \multicolumn{1}{c|}{window} &
  \multicolumn{1}{c|}{door} \\ \hline
\multicolumn{1}{|c|}{9} &
  \multicolumn{1}{c|}{10} &
  \multicolumn{1}{c|}{11} &
  \multicolumn{1}{c|}{12} &
  \multicolumn{1}{c|}{13} &
   &
   &
   \\ \cline{1-5}
\multicolumn{1}{|c|}{plant} &
  \multicolumn{1}{c|}{appliance} &
  \multicolumn{1}{c|}{furniture} &
  \multicolumn{1}{c|}{objects} &
  \multicolumn{1}{c|}{other} &
   &
   &
   \\ \hline
\multicolumn{8}{|c|}{\cellcolor[HTML]{C0C0C0}Industrial scenes} \\ \hline
\multicolumn{1}{|c|}{1} &
  \multicolumn{1}{c|}{2} &
  \multicolumn{1}{c|}{3} &
  \multicolumn{1}{c|}{4} &
  \multicolumn{1}{c|}{5} &
  \multicolumn{1}{c|}{6} &
  \multicolumn{1}{c|}{7} &
  \multicolumn{1}{c|}{8} \\ \hline
\multicolumn{1}{|c|}{pedestrian} &
  \multicolumn{1}{c|}{floor} &
  \multicolumn{1}{c|}{wall} &
  \multicolumn{1}{c|}{conveyor} &
  \multicolumn{1}{c|}{static objects} &
  \multicolumn{1}{c|}{dynamic objects} &
  \multicolumn{1}{c|}{robot} &
  \multicolumn{1}{c|}{others} \\ \hline
\multicolumn{8}{|c|}{\cellcolor[HTML]{C0C0C0}Outdoor scenes} \\ \hline
\multicolumn{1}{|c|}{1} &
  \multicolumn{1}{c|}{2} &
  \multicolumn{1}{c|}{3} &
  \multicolumn{1}{c|}{4} &
  \multicolumn{1}{c|}{5} &
  \multicolumn{1}{c|}{6} &
  \multicolumn{1}{c|}{7} &
  \multicolumn{1}{c|}{8} \\ \hline
\multicolumn{1}{|c|}{pedestrian} &
  \multicolumn{1}{c|}{bicycle} &
  \multicolumn{1}{c|}{vehicle} &
  \multicolumn{1}{c|}{road} &
  \multicolumn{1}{c|}{curb} &
  \multicolumn{1}{c|}{building} &
  \multicolumn{1}{c|}{pole} &
  \multicolumn{1}{c|}{tree} \\ \hline
\multicolumn{1}{|c|}{9} &
  \multicolumn{1}{c|}{10} &
  \multicolumn{1}{c|}{11} &
  \multicolumn{1}{c|}{12} &
   &
   &
   &
   \\ \cline{1-4}
\multicolumn{1}{|c|}{shrub} &
  \multicolumn{1}{c|}{grass} &
  \multicolumn{1}{c|}{obstacle} &
  \multicolumn{1}{c|}{others} &
   &
   &
   &
   \\ \cline{1-4}
\end{tabular}
\end{table}

In addition to the bounding boxes of dynamic targets, we also perform point-by-point semantic annotation on the point cloud.
Our definition of semantic categories to be annotated in different scenarios can be found in the \Cref{table:annotation_semantics}.
When performing semantic annotation of point clouds, we remove the dynamic targets of each clips and then superimpose the remaining static points on multiple frames of point clouds.
Then, point-level semantic annotation is performed by referring to the projection of the point cloud on the image.

The process of obtaining the occupancy ground truth after annotation can be referred to \Cref{fig:occ_generation_pipeline}.
After obtaining the static superimposed point cloud annotation and dynamic target box, we first align the superimposed static background points to the ego coordinate system of the frame-by-frame point cloud, and then splice the superimposed dynamic foreground points into the point cloud according to the dynamic target posture of the frame.
The superimposed and stitched point cloud is directly voxelized to obtain the final occupancy ground truth.
We do not perform Poisson reconstruction on the superimposed and stitched point cloud.

\subsection{Multi-Modal Fusion Network}
\label{chap:fusion_network}
Our occupancy perception model accepts multimodal inputs, including a LiDAR point cloud and 6 pinhole camera images.
We use the Bird's Eye View (BEV) paradigm that has been widely validated and adopted in autonomous driving for feature extraction and feature fusion.
Since the robotic sensors undergo pitch and roll motions during movement, it is essential to transform the sensor data into a gravity-aligned egocentric reference frame to comply with the BEV assumption.
Specifically, we extract LiDAR and camera features through two modality-specific feature extraction branches, and then perform multi-modal feature fusion through Transformer Decoder.
The final occupancy result is predicted on the fused BEV features.
The overall model architecture can be referred to \Cref{fig:model_arch}.

\noindent\textbf{Camera Feature Encoder}
In the image feature extraction branch, we use convolutional neural networks to extract features from camera images.
The images of the 6 cameras are extracted from each image through a shared backbone.
It is worth noting that our input images are distorted pinhole camera images, and we do not perform dedistortion processing to reduce the overall latency of the system.
We made adjustments in the subsequent multimodal feature fusion part so that the distorted image results will not affect the final occupancy prediction.

\noindent\textbf{LiDAR Feature Encoder}
In the LiDAR feature extraction branch, we use PointPillar~\citep{pointpillar} to construct BEV features.
The LiDAR point cloud is first voxelized into a set of non-empty pillars.
Each pillar aggregates raw points and encodes their attributes such as coordinates and reflectance via a simplified PointNet-style MLP to extract pillar-wise features.
These features are then scattered back to their original pillar locations to form a sparse pseudo-BEV feature map.
A 2D convolutional backbone further processes the sparse features into a dense BEV representation, capturing multi-scale spatial context for downstream fusion.

\noindent\textbf{Multimodal Fusion}
When performing multimodal feature fusion, we mainly refer to the DeepFusion~\citep{DeepFusion} approach.
We perform multimodal fusion through LiDAR to camera cross attention.
The LiDAR BEV features extracted by PointPillar are used as queries, and the 6 cameras convolution features are used as keys and values.
In order to more accurately describe the projection relationship from LiDAR to camera, we use Deformable Attention~\citep{DeformableDETR} for feature sampling.
When sampling camera features from LiDAR queries reference points projection, we use pinhole camera distortion projection to reduce or eliminate the impact of distorted image feature extraction.
LiDAR features provide strong geometric information, and camera features supplement semantic information.
The fused features have rich geometric and semantic features.

\noindent\textbf{Historical Feature Fusion}
To leverage temporal information while maintaining computational efficiency, we adopt a concise yet effective approach for fusing historical BEV features.
Following the paradigm established in BEVDet4D~\citep{BEVDet4D}, we maintain a dynamic feature queue storing BEV features from previous timesteps.
For each historical BEV feature map,
\begin{equation}
    F_{t-\Delta t}\in \mathbb{R}^{H\times W\times C}
\end{equation}
where $\Delta t$ represents the time difference, we first align it to the current ego coordinate frame using the relative ego-motion transformation $T_{t \rightarrow (t- \Delta t)}$:
\begin{equation}
    F^{aligned}_{t-\Delta t} = \mathcal{W} (F_{t- \Delta t, T_{t \rightarrow (t-\Delta t)}})
\end{equation}
where $\mathcal{W}(\cdot)$ denotes the bilinear interpolation-based warping operation that transforms features according to the ego-motion.
The aligned historical features are then concatenated with the current frame's BEV features along the channel dimension:
\begin{equation}
    F_t^{temp}=\mathrm{Concat}([F_t, F^{aligned}_{t-1}, \dots, F_{t-k}^{aligned}])
\end{equation}
To effectively fuse these multi-temporal features, we employ BEV encoder to further encodes the feature in the BEV space.
We utilize ResNet with classical residual block to construct the backbone and combine the features with different resolutions by applying FPN~\citep{FPN}.

\noindent\textbf{Prediction Head}
Following FlashOcc's~\citep{yu2023flashocc} efficient design, we transform 2D BEV features into 3D voxel space through channel-to-height redistribution.
FPN features are upsampled and merged through concatenation and convolution.
The consolidated BEV feature map is reshaped into 3D voxels by splitting channels into predefined height bins.
Finally, a 3D convolution block processes the 3D voxel features with minimal computational cost.
To train the proposed baselines, focal loss $\mathcal{L}_{focal}$ and lovasz-softmax loss $\mathcal{L}_{ls}$ are leveraged to optimize the network. 
Following Monoscene, we also utilize affinity loss $\mathcal{L}^{geo}_{scal}$ and $\mathcal{L}^{sem}_{scal}$ to optimize the scene-wise and class-wise metrics (i.e., geometric IoU and semantic mIoU).
Therefore, the overall loss function can be derived as:
\begin{equation}
    \mathcal{L}_{total}=\mathcal{L}_{focal} + \mathcal{L}_{ls} + \mathcal{L}^{geo}_{scal}+\mathcal{L}^{sem}_{scal}
\end{equation}

\section{Experiments and Results}

\subsection{Implementation details}
\noindent\textbf{Dataset}
Our experiments utilize a multi-modal dataset collected from 6 cameras and 1 LiDAR sensor, comprising 180 clips as training set and 20 clips as evalidation set.
Each clip contains 200 frames, ensuring rich temporal context for dynamic scene understanding.
The LiDAR data provides 3D bounding box annotations and point-wise semantic segmentation labels, which are leveraged to generate high-quality voxelized occupancy ground truth.

\noindent\textbf{Metrics}
We evaluate our occupancy predictions using two metrics.
We use mIoU to measure semantic segmentation accuracy in 3D space, computed as the mean IoU across all classes.
Following SparseOcc~\citep{liu2023fully}, we use rayIoU by computing IoU along sampled lidar rays to solve the inconsistency penalty along the depth axis.

\noindent\textbf{Network settings}
The occupancy prediction range is set to [-10m, 10m] for the X and Y axes, and [-1.5m, 0.9m] for the Z-axis, all within the ego coordinate system.
The final output 3D occupancy grid has a shape of 200×200×24, with each voxel measuring [0.1m, 0.1m, 0.1m].
For the camera branch, we adopt a pretrained ResNet50 with FPN as the 2D backbone, where input multi-view images are resized to $960 \times 768$.

\noindent\textbf{Training details}
Our model is implemented based on mmdetection3d framework and trained on 8 NVIDIA A100 GPUs with a batch size of 4.
We adopt the AdamW optimizer with a cosine annealing learning rate scheduler, incorporating a warmup phase, and set the initial learning rate to 2e-4.
The model is trained for 20 epochs using a StreamPETR-style~\citep{wang2023exploring} continuous frame training strategy and temporal information is incorporated starting from the second epoch.

\subsection{Main results}
\begin{figure}[]
    \centering
    \includegraphics[width=0.8\textwidth]{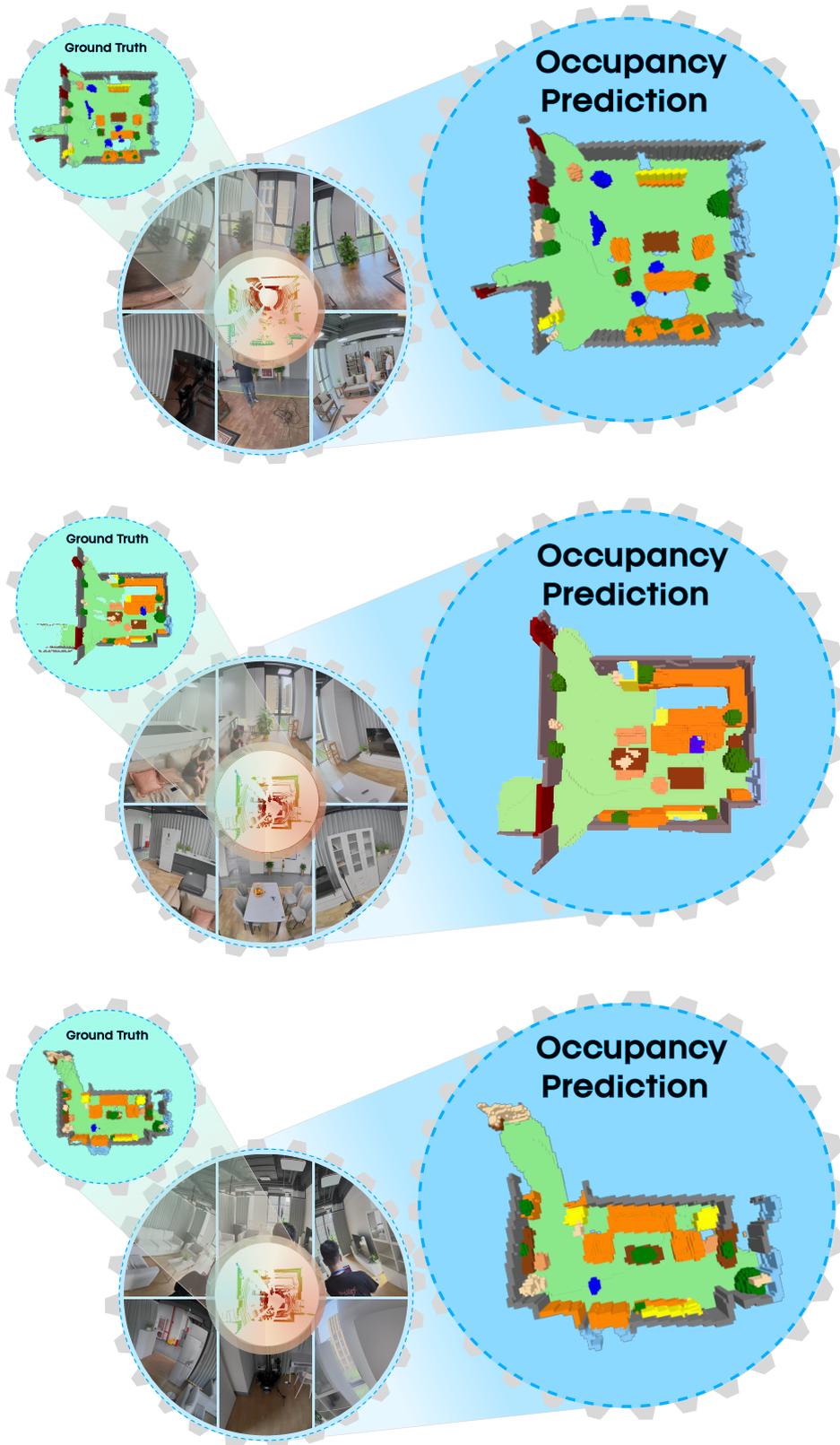}   
    \caption{\textbf{The results of our humanoid perception system.} The left side of each row shows the occupancy ground truth, the middle is the 6 images and point cloud input to the model, and the right side shows the result of model inference.}
    \label{fig:perception_results}
\end{figure}
\begin{figure}[]
    \centering
    \begin{subfigure}[b]{0.48\textwidth}
        \centering
        \includegraphics[width=\textwidth]{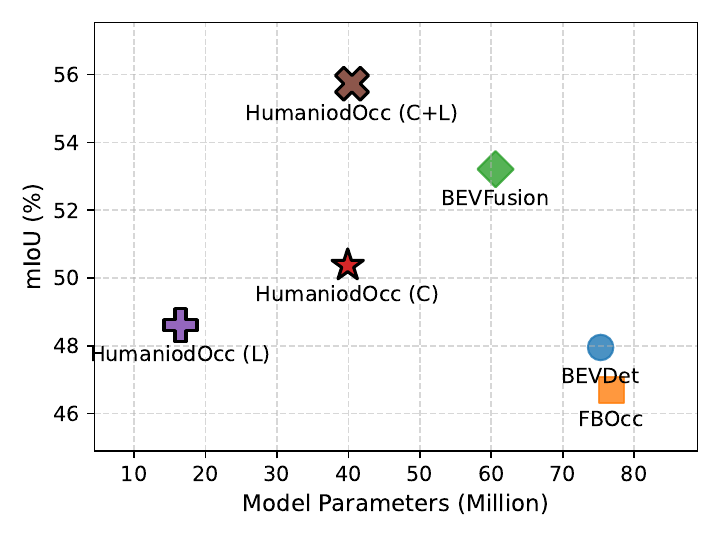}
        \caption{mIoU performance}
    \end{subfigure}
    \hfill
    \begin{subfigure}[b]{0.48\textwidth}
        \centering
        \includegraphics[width=\textwidth]{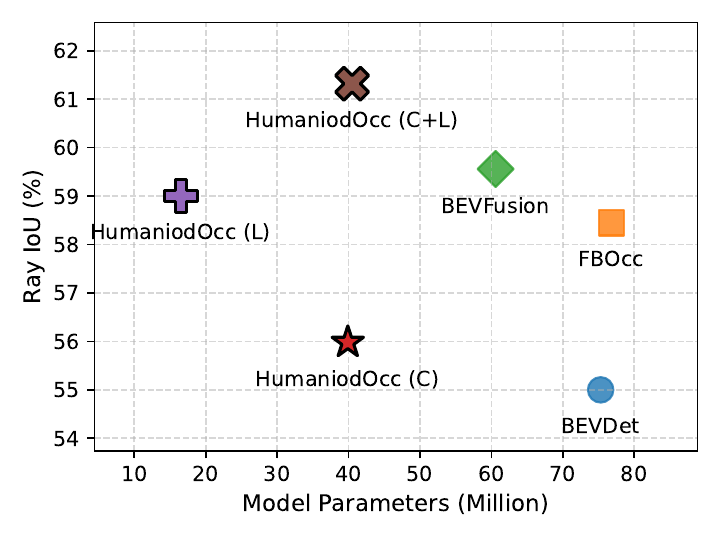}
        \caption{ray IoU performance}
    \end{subfigure}

    \caption{\textbf{Comparisons of the mIoU and ray IoU of various 3D
occupancy prediction methods.} Our model significantly outperforms other methods in both mIoU and ray IoU with the smallest number of parameters.}
    \label{fig:model_performance}
\end{figure}

We benchmark our method against representative BEV perception models on our multi-modal dataset.
Our comparative analysis evaluates occupancy prediction performance both on single-frame settings and multi-frame settings with historical information.
The comparison between our model and some general methods can be seen in~\Cref{fig:model_performance}.
All models adopt identical training configurations, including input image resolution, backbone network, feature dimensions, and training strategies.
As shown in \Cref{table:main_results}, our model achieves superior metrics while maintaining lightweight architecture with significantly fewer parameters.
We also show the visualization results of our perception system in \Cref{fig:perception_results}.

\begin{table}[]
\renewcommand\arraystretch{1.2}
\caption{\textbf{3D semantic occupancy prediction performance on our dataset.} For a fair comparison,
we train all models on our dataset with same setups.}
\centering
\resizebox{\textwidth}{!}{
\begin{tabular}{l|cc|c|cc|cccccccccccc}
\toprule
\multicolumn{1}{c}{\multirow{2}{*}{Method}} & \multicolumn{2}{c}{Settings}                               & \multicolumn{1}{c}{\multirow{2}{*}{Params}} & \multicolumn{2}{c}{Metrics}                            & \multicolumn{12}{c}{Class Metrics}                                                                                                                                                                                                                                                                                                                                     \\ \cline{2-3} \cline{5-18} 
\multicolumn{1}{c}{}                        & \multicolumn{1}{c}{Modality} & \multicolumn{1}{c}{Frames} & \multicolumn{1}{c}{}                        & \multicolumn{1}{c}{mIoU} & \multicolumn{1}{c}{rayIoU} & \multicolumn{1}{c}{pedestrian} & \multicolumn{1}{c}{robot} & \multicolumn{1}{c}{chair} & \multicolumn{1}{c}{table} & \multicolumn{1}{c}{floor} & \multicolumn{1}{c}{wall} & \multicolumn{1}{c}{window} & \multicolumn{1}{c}{door} & \multicolumn{1}{c}{plant} & \multicolumn{1}{c}{appliance} & \multicolumn{1}{c}{furniture} & \multicolumn{1}{c}{objects} \\ \hline
BEVDet~\citep{huang2021bevdet}                                        & C                             & 1                          & 75.3M                                        & 47.93                     & 59.13                       & 39.85                           & 55.93                      & 43.31                      & 47.76                      & 60.79                      & 51.91                     & 37.77                       & 44.27                     & 60.25                      & 50.89                          & 54.76                          & 27.66                        \\
FBOcc~\citep{li2023fb}                                         & C                             & 1                          & 76.8M                                        & 47.37                     & 59.10                        & 39.78                           & 55.43                      & 42.95                      & 45.37                      & 62.69                      & 51.19                     & 35.73                       & 44.57                     & 59.12                      & 49.36                          & 53.99                          & 28.27                        \\
BEVFusion~\citep{liu2022bevfusion}                                     & C+L                           & 1                          & 60.6M                                        & \cellcolor{gray!50}53.98                     & 60.24                       & 51.00                              & 51.70                       & 36.72                      & 43.31                      & 66.64                      & 63.38                     & 41.42                       & 71.40                      & 65.29                      & 61.38                          & 62.96                          & 32.45                        \\
\textbf{HumaniodOcc~(Ours)}                                   & C+L                           & 1                          & \cellcolor{gray!50}40.5M                                        & \cellcolor{gray!20}52.79                     & \cellcolor{gray!50}60.49                       & 49.23                           & 49.22                      & 35.28                      & 43.63                      & 57.66                      & 63.90                      & 43.07                       & 71.41                     & 64.17                      & 60.89                          & 59.43                          & 35.54                        \\
\hline
BEVDet~\citep{huang2021bevdet}                                        & C                             & 2                          & 75.3M                                        & 47.95                     & 55.00                          & 39.37                           & 55.66                      & 28.32                      & 39.29                      & 62.42                      & 51.56                     & 36.85                       & 58.61                     & 63.85                      & 54.96                          & 55.03                          & 29.52                        \\
FBOcc~\citep{li2023fb}                                         & C                             & 2                          & 76.8M                                        & 46.70                      & 58.45                       & 40.11                           & 53.52                      & 42.97                      & 45.44                      & 62.14                      & 50.83                     & 35.38                       & 41.06                     & 58.08                      & 49.07                          & 54.00                             & 27.76                        \\
BEVFusion~\citep{liu2022bevfusion}                                     & C+L                           & 2                          & 60.6M                                        & 53.21                     & 59.56                       & 51.00                              & 52.16                      & 37.03                      & 40.83                      & 57.63                      & 63.88                     & 44.38                       & 71.68                     & 63.87                      & 62.45                          & 61.16                          & 32.43                        \\
\textbf{HumaniodOcc~(Ours)}                                   & C+L                           & 2                          & \cellcolor{gray!50}40.5M                                        & \cellcolor{gray!50}55.73                     & \cellcolor{gray!50}61.32                       & 50.85                           & 56.16                      & 40.49                      & 45.49                      & 65.60                       & 64.76                     & 44.07                       & 72.77                     & 65.58                      & 63.01                          & 64.50                           & 35.48                       \\
\bottomrule
\end{tabular}}
\label{table:main_results}
\end{table}

\subsection{Ablation study}
Ablation studies are conducted to validate the effectiveness of components in our approach.
We analyze the impacts of camera distortion correction strategies, temporal multi-frame aggregation, and different sensor modality inputs  on occupancy prediction performance.

\noindent\textbf{Ablation of distortion}
\begin{table}[]
\centering
\caption{\textbf{Comparisons of the mIoU and ray IoU of the proposed distortion processing methods.} The method we finally implemented avoids undistorting the original image to retain the most effective pixels of the image. Sampling by distorted projection ensures that the performance is not affected.}
\label{table:distort}
\begin{tabular}{l|cc|cc}
\toprule
Method      & Input images & Projection & mIoU  & rayIoU \\
\hline
HumaniodOcc & raw          & pinhole   & 46.23 & 52.54  \\
HumaniodOcc & undistorted  & pinhole   & 47.41 & 53.46  \\
\rowcolor{gray!50}HumaniodOcc & raw          & distort    & 47.92 & 53.97 \\
\bottomrule
\end{tabular}
\end{table}
As mentioned in \Cref{chap:fusion_network}, we evaluate three distortion handling strategies for images in \Cref{table:distort}.
Using raw distorted images directly achieves 46.23 mIoU but introduces geometric errors in BEV space.
Traditional image undistortion preprocessing improves accuracy to 47.41 mIoU but incurs prohibitive latency for robotic deployment scenarios. Our proposed distortion-aware projection method demonstrates superior performance of 47.92 mIoU by encoding lens distortion parameters directly into view transformation.
Notably, for this comparison we use data with precisely calibrated projection alignment to reveal the distinct characteristics of each distortion handling method in terms of geometric accuracy.

\noindent\textbf{Ablation of temporal}
\begin{table}[]
\centering
\caption{\textbf{Comparisons of the mIoU and ray IoU of the different number of temporal frames.} Using 2 frames of temporal information can achieve the best model performance.}
\label{table:temporal}
\begin{tabular}{l|c|cc}
\toprule
Method      & Frames & mIoU  & rayIoU \\
\hline
HumaniodOcc & 1     & 52.79 & 60.49  \\
\rowcolor{gray!50}HumaniodOcc & 2     & 55.73 & 61.32  \\
HumaniodOcc & 3     & 55.11 & 60.53  \\
HumaniodOcc & 4     & 54.3  & 60.63 \\
\bottomrule
\end{tabular}
\end{table}
We conduct an ablation study on the number of aggregated frames in \Cref{table:temporal}.
The temporal fusion module effectively aggregates historical frame features, which yields significant improvements over single frame model, primarily by enhancing motion awareness and occulsion reasoning.
However, incorporating more than one historical frame leads to performance degradation, which we attribute to the pose errors accumulation during robotic motions.

\noindent\textbf{Ablation of modalities}
\begin{table}[]
\centering
\caption{\textbf{Comparisons of the mIoU and ray IoU of the different number of modalities.} The model can obtain good occupancy prediction results when using only LiDAR information, but the semantic differentiation is relatively poor. The semantic results of the model using only camera information are better than using only LiDAR, but the occupancy prediction is relatively poor due to the lack of clear geometric information in the image. Our fusion method retains the LiDAR geometric information and obtains rich semantic information from the camera.}
\label{table:modalities}
\begin{tabular}{l|c|cc}
\toprule
Method      & Modal & mIoU  & rayIoU \\
\hline
HumaniodOcc & C     & 50.37 & 55.98  \\
HumaniodOcc & L     & 48.61 & 59.01  \\
\rowcolor{gray!50}HumaniodOcc & C+L   & 55.73 & 61.32  \\
\bottomrule
\end{tabular}
\end{table}
\Cref{table:modalities} demonstrates the expected performance hierarchy: camera-only (50.37 mIoU), LiDAR-only (48.61 mIoU), and multi-modal fusion (55.73 mIoU).
The improvement of fusion model confirms the complementary strengths of both modalities - LiDAR provides precise depth while cameras offer rich semantics.
This validates our architecture's effective cross-modal fusion capability.
\section{Conclusion and Future Work}

\begin{wrapfigure}{r}{\textwidth}
    \centering
    \includegraphics[width=1.0\linewidth]{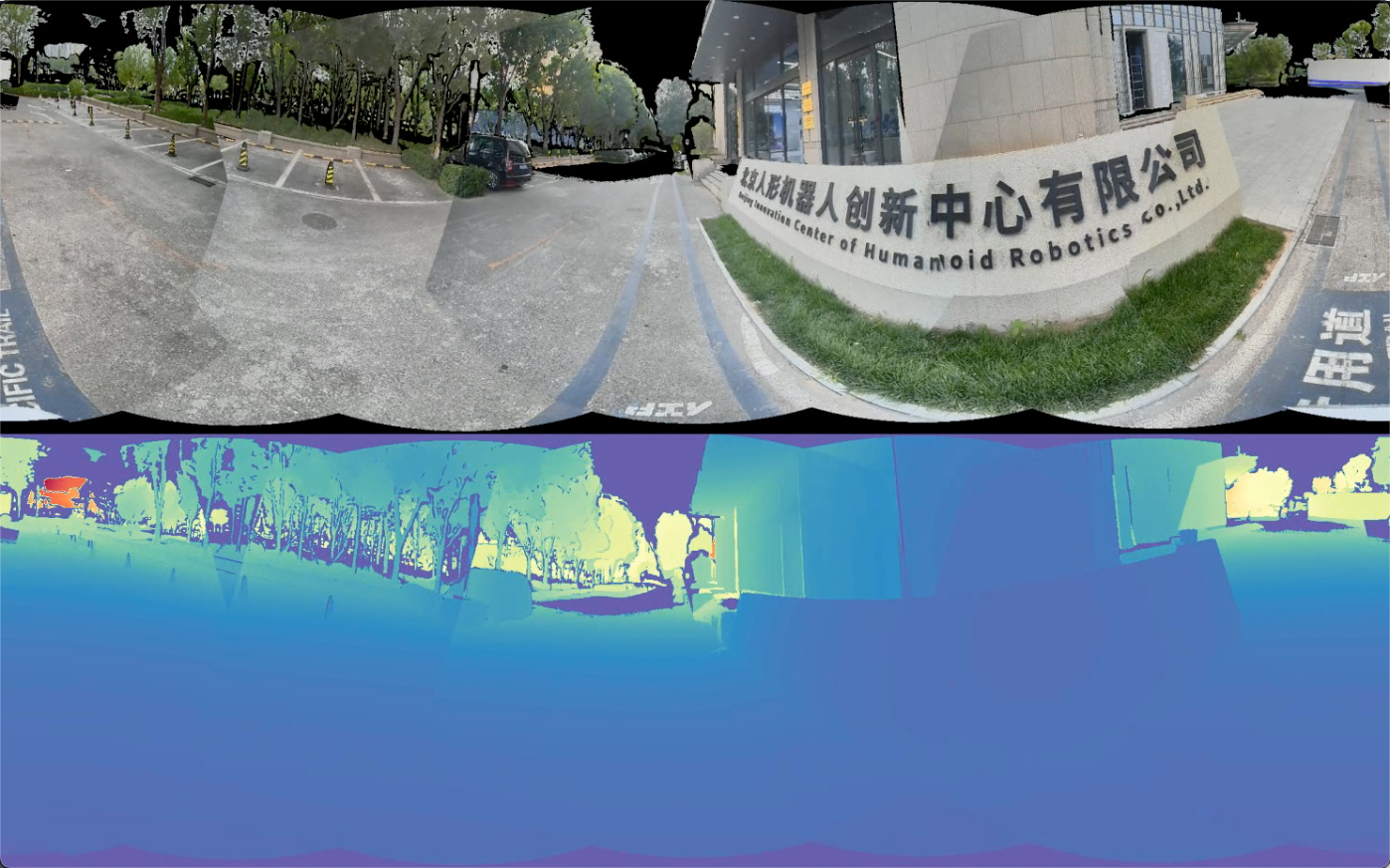}
    \caption{Panoramic stitching of 6 cameras and densified visualization of LiDAR point clouds.}
    \label{fig:depth_camera}
\end{wrapfigure}

\begin{wrapfigure}{r}{\textwidth}
    \centering
    \includegraphics[width=1.0\linewidth]{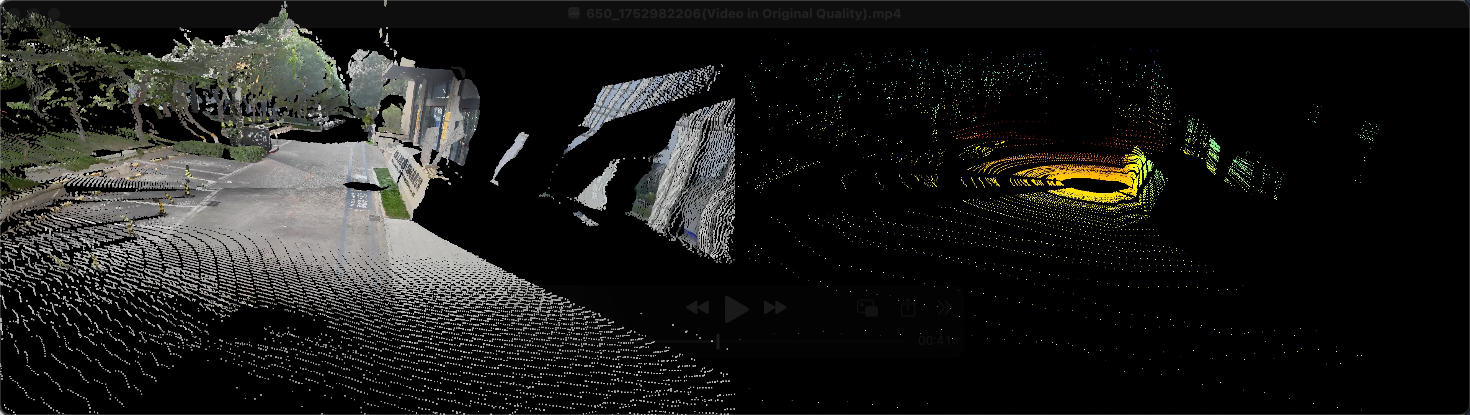}
    \caption{3D scene visualization and LiDAR point clouds visualization.}
    \label{fig:lidar_camera}
\end{wrapfigure}

In this work, we introduced Humanoid Occupancy, a unified multimodal perception system tailored for humanoid robots. By leveraging occupancy-based representations, our framework enables rich semantic and geometric understanding of complex environments, and addresses key challenges in sensor layout, data annotation, and multimodal fusion. We constructed the first panoramic occupancy dataset for humanoid robots and demonstrated the effectiveness of the system on the Tienkung platform in real-world scenarios. Looking ahead, we recognize that omnidirectional perception and mapping are crucial for the future of remote robot control. We plan to utilize state-of-the-art computer vision reconstruction techniques in combination with occupancy representations to build more efficient omnidirectional perception systems~(\Cref{fig:depth_camera} and \Cref{fig:lidar_camera}), further expanding the dataset, refining temporal fusion strategies, and exploring broader deployment across diverse humanoid platforms to advance robust and standardized visual perception in robotics.

\section*{Acknowledgements}

We sincerely thank Hongdi Li and Dengke Shang for their outstanding support and guidance as project managers during the development of this work, and appreciate the dedicated efforts of the X-Humanoid engineering team (Junjie Hu, Qiong Wu, Jian Xiao, Chenghao Sun, Hengle Ren, Wenjing Deng, Hao Yang, Shuo Xu, Dawei Gao, Fachao Zhang, Yuqing Dong) in driving the progress of the Humanoid Occupancy exploratory project. We are especially grateful to the leadership team (Youjun Xiong, Chunzhi Li and Jian Tang) of X-Humanoid for their strategic vision, continuous encouragement, and strong support throughout the project. We also thank Ruiying Zhang for helping with the production of \Cref{fig:perception_results}, and Gabriele Meilikhov for his contributions to the visual presentation.


\clearpage
\setcitestyle{numbers}
\bibliographystyle{plainnat}
\bibliography{main}

\end{document}